\documentclass[runningheads]{llncs}
\usepackage{graphicx}
\usepackage{cite} 
\usepackage{times}
\usepackage{epsfig}
\usepackage{amsmath}
\usepackage{amssymb}
\usepackage{mwe}
\usepackage{acro}
\usepackage{amssymb}
\usepackage{xcolor,colortbl}
\usepackage{tabularx}
\usepackage{relsize}
\usepackage{pifont}
\usepackage{booktabs} 
\usepackage{multirow}
\usepackage{multicol}
\usepackage{adjustbox}
\usepackage{float}

\usepackage{cite}
\usepackage{amsmath,amssymb,amsfonts}
\usepackage{graphicx}
\usepackage{textcomp}

\usepackage{algorithmicx}               
\usepackage{algorithm} 
\usepackage{algpseudocode}  

\usepackage{array}
\usepackage{booktabs}
\usepackage{colortbl}
\definecolor{dark-gray}{gray}{0.7} 
\usepackage{multirow}

\usepackage{float}
\usepackage{cuted}
\usepackage{graphicx, caption}
\usepackage[colorlinks,
            linkcolor=red,  
            anchorcolor=blue, 
            citecolor=green,   
            ]{hyperref}

\DeclareAcronym{ROI}{
short=ROI,
long=region of interest,
}

\DeclareAcronym{IOU}{
short=IOU,
long=intersection over union,
}

\DeclareAcronym{cIOU}{
short=cIOU,
long=circle intersection over union,
}

\DeclareAcronym{DoF}{
short=DoF,
long=degrees of freedom,
}

\begin{document}
\title{VoxelEmbed: 3D Instance Segmentation and Tracking with Voxel Embedding based Deep Learning}
%
%
\author{Mengyang Zhao\inst{1} \and
Quan Liu \inst{2} \and
Aadarsh Jha \inst{2} \and
Ruining Deng \inst{2} \and
Tianyuan Yao \inst{2} \and
Anita Mahadevan-Jansen \inst{2} \and
Matthew J.Tyska \inst{2} \and
Bryan A. Millis \inst{2} \and
Yuankai Huo\inst{2}}


%
 \institute{******}
\institute{
Dartmouth College, Hanover, NH 03755 \and
Vanderbilt University, Nashville TN 37215, USA 
}


%
\maketitle              
\begin{abstract}
Recent advances in bioimaging have provided scientists a superior high spatial-temporal resolution to observe dynamics of living cells as 3D volumetric videos. Unfortunately, the 3D biomedical video analysis is lagging, impeded by resource insensitive human curation using off-the-shelf 3D analytic tools. Herein, biologists often need to discard a considerable amount of rich 3D spatial information by compromising on 2D analysis via maximum intensity projection. Recently, pixel embedding based cell instance segmentation and tracking provided a neat and generalizable computing paradigm for understanding cellular dynamics.  In this work, we propose a novel spatial-temporal voxel-embedding (VoxelEmbed) based learning method to perform simultaneous cell instance segmenting and tracking on 3D volumetric video sequences. Our contribution is in four-fold: (1) The proposed voxel embedding generalizes the pixel embedding with 3D context information; (2) Present a simple multi-stream learning approach that allows effective spatial-temporal embedding; (3) Accomplished an end-to-end framework for one-stage 3D cell instance segmentation and tracking without heavy parameter tuning; (4) The proposed 3D quantification is memory efficient via a single GPU with 12 GB memory. We evaluate our VoxelEmbed method on four 3D datasets (with different cell types) from the ISBI Cell Tracking Challenge. The proposed VoxelEmbed method achieved consistent superior overall performance (OP) on two densely annotated datasets. The performance is also competitive on two sparsely annotated cohorts with 20.6$\%$ and 2$\%$ of data-set having segmentation annotations. The results demonstrate that the VoxelEmbed method is a generalizable and memory-efficient solution.
\end{abstract}
\section{Introduction}
Characterizing cellular dynamics, the behaviors of the fundamental units of life, is indispensable in translational biological research, such as organogenesis~\cite{cao2019single}, immune response~\cite{ong2020dynamic}, drug development~\cite{zhou2006high}, and cancer metastasis~\cite{condeelis2006macrophages}. With a tenet of “seeing is believing”, recent advances in bioimaging have provided scientists unprecedented high spatial-temporal resolution to observe three dimensional dynamics (3D volumes + time) of living cells~\cite{liu2018observing}. Yet, large-scale bioimage data are concurrent with imaging innovations, causing fundamental computational challenges for quantifying cellular dynamics in translational biological research, where “quantifying is deciding”~\cite{meijering2020bird}.  For example, a single lattice light-sheet microscope~\cite{chen2014lattice} (LLS) produce TB level rich spatial-temporal dynamic 3D volumetric videos~\cite{wan2019light}. Unfortunately, the 3D biomedical video analysis is lagging, impeded by resource insensitive human curation via off-the-shelf 3D analytic tools~\cite{von2019artificial}. Recent deep learning techniques have achieved remarkable success in computer vision and biomedical image analysis. However, the large-scale quantification of “3D+time” cellular dynamics with deep learning (e.g., dense instance segmentation and tracking) is still hindered by the high dimensionality and heterogeneity of the dynamics.~\cite{jiang2016enhanced} Herein, biologists often need to discard a considerable amount of rich 3D spatial information by projecting the 3D videos to 2D space for downstream analyses.

\begin{figure}[t]
\begin{center}
\includegraphics[width=3 in]{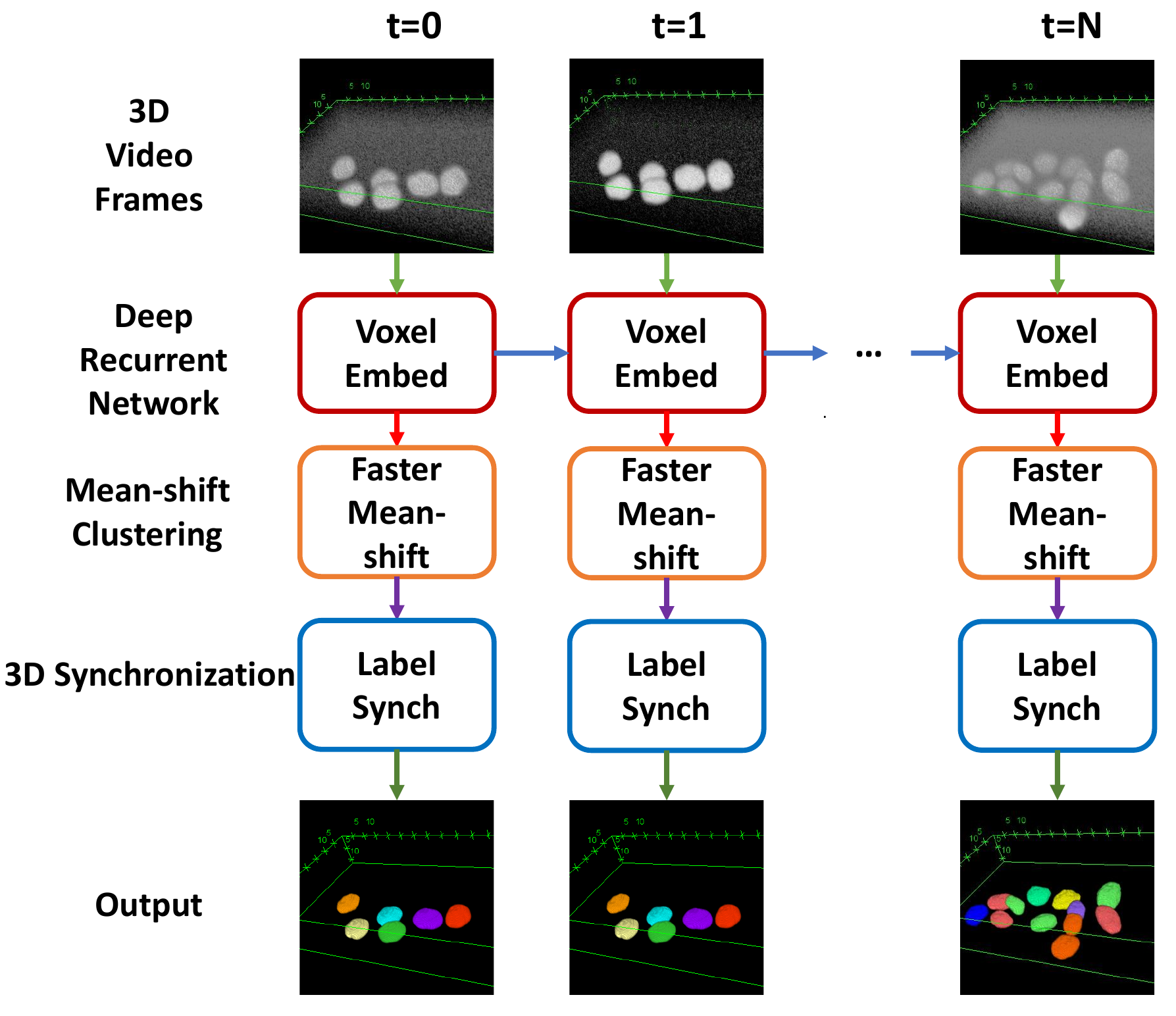}
\end{center}
   \caption{\textbf{The overall framework.} The workflow of the proposed voxel embedding based deep learning framework is presented, for 3D cell instance segmentation and tracking.}
\label{fig2}
\end{figure}

When quantifying cellular dynamics, the “segment-then-track” two-stage paradigm ~\cite{debeir2005tracking,jin2020deep,condeelis2006macrophages,zhao2020cloud} is a prevalent design for both conventional model based methods and deep learning approaches. Such a paradigm first segments instance objects across frames and then links the instance objects via association algorithms. However, the spatial and temporal information from the same individual object cannot be learned simultaneously with the two-stage design. To integrate and handle these two tasks simultaneously, Payer et al.~\cite{payer2019segmenting} proposed a cosine pixel embedding based recurrent stacked hourglass network (RSHN) for simultaneous instance cell segmentation and tracking. The pixel embedding approach tackled instance segmentation and tracking within an uniformed ``single-stage" framework. The key idea is to loosen the constraint of having each cell requiring a globally unique embedding, to just allowing the cell to have a different embedding relative to the nearby four cells (based on the Four Color Map theorem~\cite{appel1976every,payer2019segmenting}). Herein, the number of embeddings does not have to strictly increase with the number of cells, providing a scalable learning strategy. Based on the merits, it is appealing to adapt such strategy from 2D to 3D settings. However, the direct 3D adaptation (e.g., change a 2D network to a 3D version) requires higher computational resources, more training samples, and a considerably larger embedding space to differentiate each cell to all of its neighbors in 3D.

In this study, we propose a novel spatial-temporal voxel-embedding (VoxelEmbed) deep learning based method to perform 3D cell instance segmentation and tracking. As opposed to deploying a 3D network, we develop a simple multi-stream learning approach to learn spatial, temporal, and 3D context information simultaneously via a 2D network design. To aggregate 3D information, we introduce a 3D synchronization algorithm to build volumetric masks inspired by recent advances in slice propagation~\cite{cai2018accurate}. Briefly, the innovations of the proposed approach is in four-folds: 

(1) The proposed VoxelEmbed approach generalizes the pixel embedding to a voxel embedding with 3D context information. 
	
(2) A simple multi-stream learning approach is presented that allows effective spatial-temporal embedding.
	
(3) To our knowledge, this is the first embedding based deep learning approach for simultaneous 3D cell instance segmentation and tracking.
	
(4) The proposed method is memory efficient to a single GPU with 12 GB memory. 

\section{Methods}
The principle of our proposed VoxelEmbed framework is presented Fig. \ref{fig2}. In VoxelEmbed, we extend spatial-temporal embedding by adding 3D information. The extra embedding information is obtained from the zigzag multi-stream straining.

\subsection{Cosine Embedding based Instance Segmentation and Tracking} Payer et al.~\cite{payer2019segmenting} proposed a cosine embedding based recurrent stacked hourglass network (RSHN) that, for the first time, integrated the instance segmentation and tracking algorithms into a one-stage holistic learning framework with pixel-wise cosine embedding, which achieves instance cell segmentation and tracking in a one-stage framework. The entire pixel embedding based learning consist of two major stages: embedding encoding and feature clustering. 

The RSHN network, combining convolutional GRUs~\cite{ballas2015delving,jiang2020automatic} (ConvGRUs)  and the stacked hourglass network~\cite{newell2016stacked}, is employed as the backbone network for our voxel embedding. In~\cite{payer2019segmenting}, each pixel in a 2D video sequence was encoded to a high-dimensional embedding vector with the intuition that all pixels from the same cell, across spatial and temporal, should have the same feature representation (embedding). The renown cosine similarity~\cite{li2004similarity,liu2021towards} approach is commonly used to measure the similarities of any two pixels. For instance, $\bf{A}$ was the embedding vector of pixel a, and $\bf{B}$ was the embedding vector of pixel b. The cosine similarity of $\bf{A}$ and $\bf{B}$ is defined as:

\begin{equation}
cos\left( {{\bf{A}},{\bf{B}}} \right){\rm{  = }}\frac{{{\bf{A}} \cdot {\bf{B}}}}{{\left\| {\bf{A}} \right\|{\rm{ }}\left\| {\bf{B}} \right\|}}
\label{eq1}\end{equation}which ranged from -1 to 1, where 1 indicates that two vectors have the same direction, 0 indicates orthogonal, and -1 indicates the opposite. The cosine similarities is used as a loss function to force the pixels from the same cell to have $cos\left( {{\bf{A}},{\bf{B}}} \right)$ towards $1$, while the pixels from different cells to be $0$

\subsection{Voxel Embedding (Training Stage)}
\begin{figure}[t]
\begin{center}
\includegraphics[width=4 in]{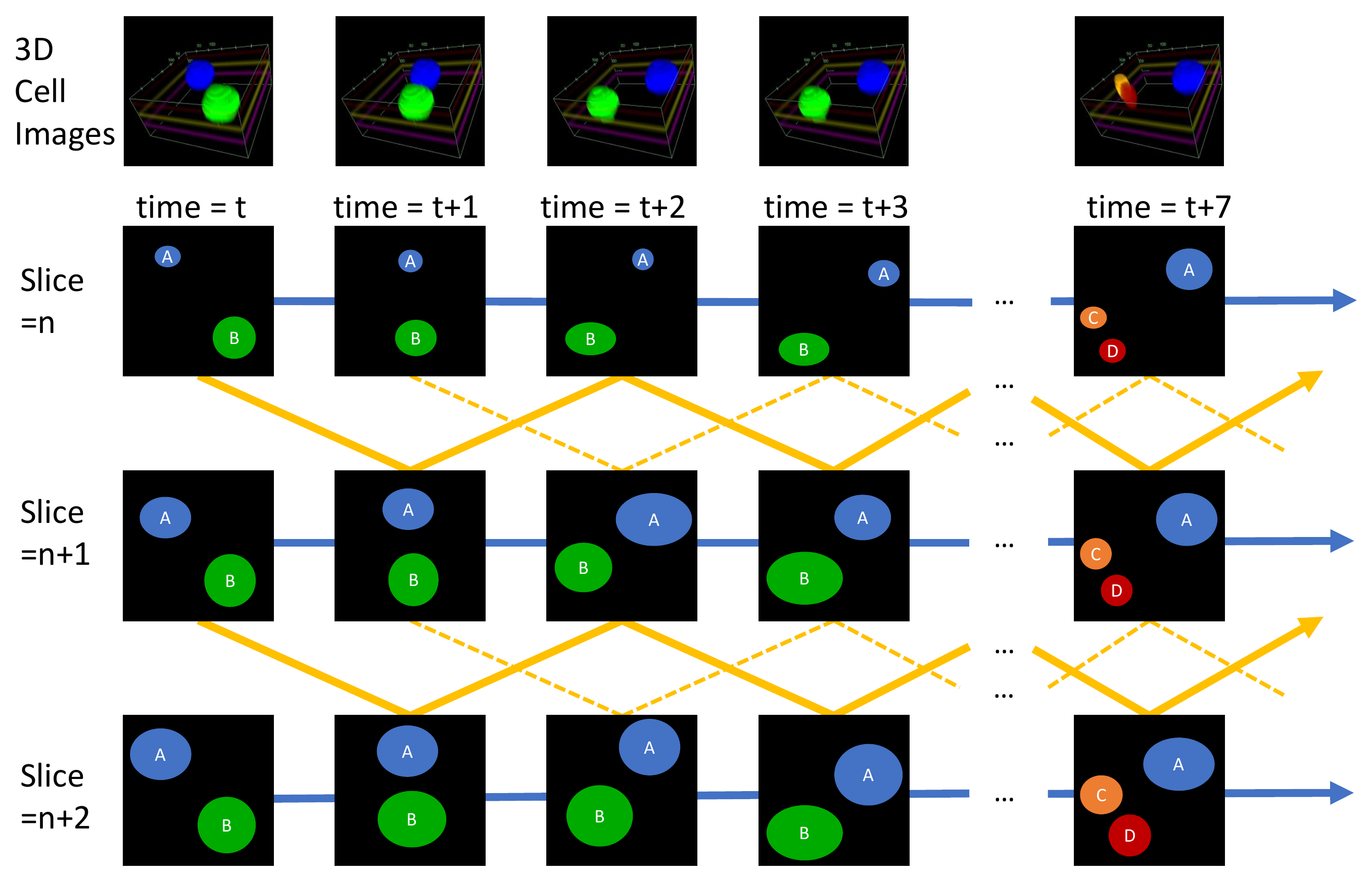}
\end{center}
\caption{\textbf{VoxelEmbed at the training stage.} Our VoxelEmbed model is trained by both blue and yellow streams, to ensure the same voxel embedding of the same object in terms of spatial-temporal domain and 3D context.}
\label{fig3}
\end{figure}

To encode the 3D context information for 3D cell instance segmentation and tracking, we generalize the pixel embedding principle to a voxel embedding scheme. Briefly, the VoxelEmbed approach learns additional 3D context features, and concatenates those features to original pixel embedding features. The additional 3D context features are learned from the new zigzag training path (the golden stream in Fig.\ref{fig3}) beyond the temporal embedding path (the blue stream in Fig. \ref{fig3}). 

The sections with the same z-axis location from all 3D volumes across the temporal direction are used as the temporal embedding path, to ensure the same embedding of each cell (at the same z-axis location in 2D) has the same embedding along with  migration and time. 

Then, the sections at nearby z-axis locations are selected in a zigzag training path to enforce the embedding similarity along with the migration and 3D context, when learning all zigzag training paths converge the entire 3D space. The zigzag training paths are designed as an interleaved ``W" shape. Using multi-stream training, the voxels from the same cell in a 3D video will be forced to have the same embedding along with migration, time, and 3D context.

\begin{figure}[t]
\begin{center}
\includegraphics[width=4 in]{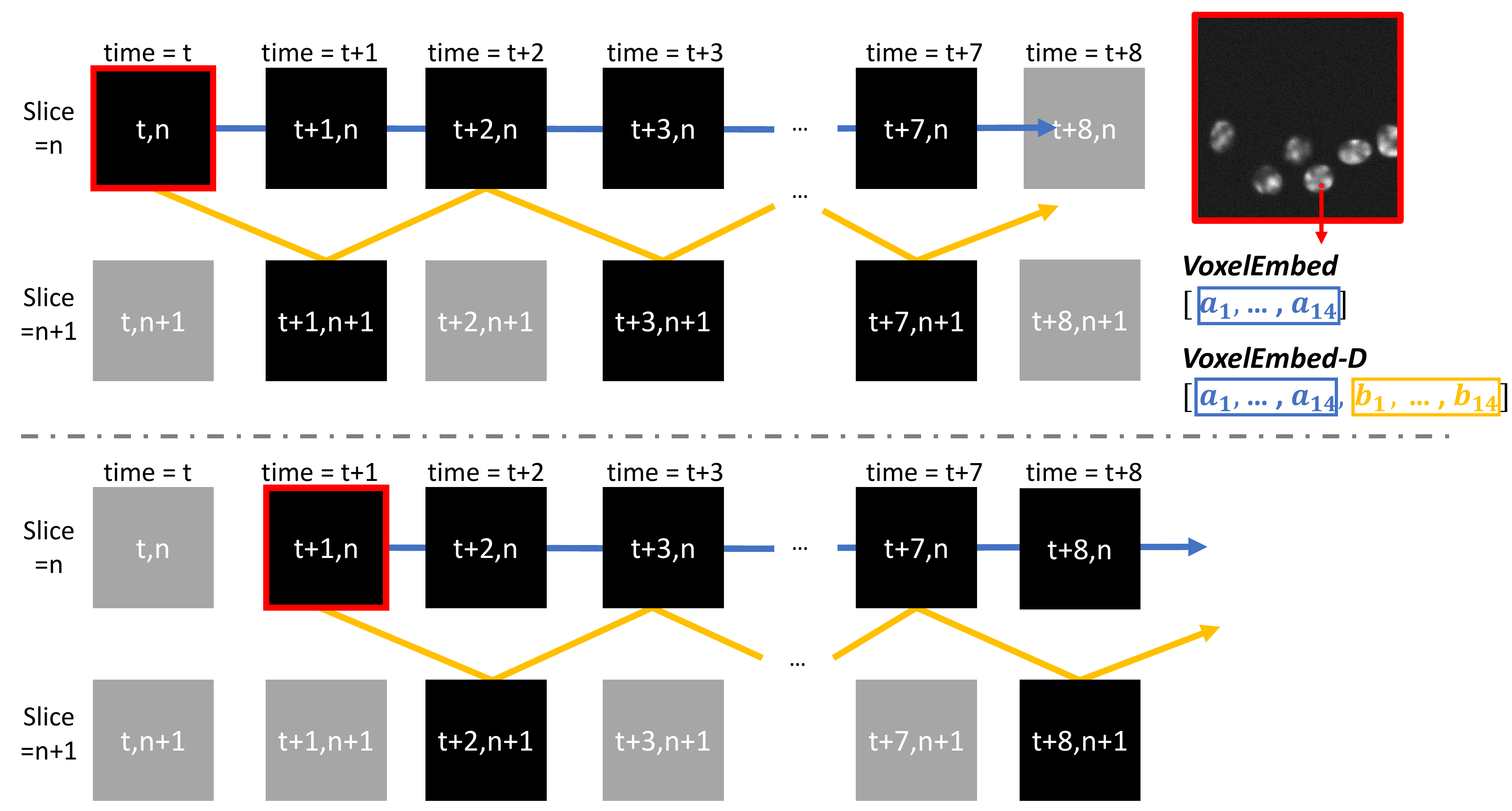}
\end{center}
\caption{\textbf{VoxelEmbed at the testing stage.} The blue and yellow streams cover two ways of forming testing samples (e.g., T = 8 in this study) for obtaining embedding of the first frame (red boarder) at the testing stage. With the embedded feature maps, the 3D cell instance segmentation and tracking results are obtained from the 3D synchronization (see \textbf{Algorithm ~\ref{alg:alg1}}).}
\label{fig4}
\end{figure}

\subsection{Voxel Embedding (Testing Stage)}

In implementation, $T$ frames from time $t$ to $t+T$ are used as a single training sample. The temporal embedding feature, a 14-dimensional feature vector, is encoded for each voxel: 
\begin{equation}
f_{time} = \left[ {{a_1},{a_2},...,{a_{14}}} \right]
\end{equation}

With the same principle, $T$ frames from time $t$ to $t+T$ with zigzag frames are used as another single training example. The 3D context embedding feature, another 14-dimensional feature vector, is encoded for each voxel:
\begin{equation}
f_{threeD} = \left[ {{b_1},{b_2},...,{b_{14}}} \right]
\end{equation}

Then, the $f_{time}$ and $f_{threeD}$ are concatenated into a 28 dimension feature for each voxel as our final voxel embedding feature:

\begin{equation}
f_{spatial-temporal} = {f_{time}} \oplus f_{threeD} = \left[ {{a_1},...,{a_{14}},{b_1},...,{b_{14}}} \right]
\end{equation} where $\oplus$ is the concatenation operator. 




In the testing stage, the encoded embedding images will be clustered to each individual cells. The standard Mean-shift clustering algorithm was used in Payer et al.~\cite{payer2019segmenting}. In this study, we utilize a GPU accelerated Faster Mean-shift algorithm~\cite{zhao2020faster} to accelerate the embedding clustering. From the unsupervised clustering, a unique label will be assigned to each cell as different instances, which achieve the instance segmentation and tracking simultaneously. 


\subsection{3D Synchronization}


In the original pixel-embedding approach, the temporal synchronization step was introduced to stick different short "video clips" to the a complete full cell video, with consistent instance numbers. Inspired by this approach, we extend the synchronization from temporal to spatial as well.  To do so, we propose a 3-D synchronization \textbf{Algorithm \ref{alg:alg1}} to improve label synchronization on the z-axis direction.

\begin{algorithm}[h]  
  \caption{3D Synchronization Algorithm}  
  \label{alg:alg1}  
  \begin{algorithmic}[1] 
    \Require 
      $InMasks$: Unsynchronized mask-set;  
      $ReMasks$: Reference mask-set; 
    \Ensure  
      $OutMasks$: Synchronized mask-set;   
    
    \For{${mask} \in InMasks$}
        \For{${rmask} \in ReMasks$}
        \State ${{S}_{mask}} \gets {J}(rmask,mask)$ \# Caluate the Jaccard similarity: $J\left( {R,S} \right) = \frac{{\left| {R \cap S} \right|}}{{\left| {R \cup S} \right|}}$
        \If{${{S}_{mask}} $ is largest in the layer of ${mask}$}
            \State ${{List}_{rmask}}$ append ${mask}$ \# Get the largest similarity instance in each layers
        \EndIf
        \EndFor
    \EndFor
    \For{${list} \in {{List}_{rmask}}$}
        \State Find the most common label $l \in {list}$ 
        \For{${mask} \in {list}$}
            \State ${mask}\gets l$ \# Synchronize the mask
        \EndFor
    \EndFor
  \end{algorithmic}  
\end{algorithm} 

In \textbf{Algorithm \ref{alg:alg1}}, the input reference mask-set $ReMasks$ is the layer with the largest foreground ratio. In Jaccard Similarity~\cite{niwattanakul2013using}, $R$ is the set of pixels belonging to reference mask and $S$ is the set of pixels belonging to matching mask


\section{Data and Implementation Details}
In this study, we conduct an empirical validation via the ISBI Cell Tracking Challenge~\cite{ulman2017objective} dataset to evaluate the accuracy performance of our proposed framework. Specifically, we used 3-D microscope video sequences from the ISBI Cell Tracking Challenge, which is independent with training data. The following four video sequence datasets of different sizes, shapes, and textures cells were adopted to evaluate the performance: (1) Chinese Hamster Ovarian nuclei(Fluo-N3DH-CHO), (2) C.elegans developing embryo (Fluo-N3DH-CE), (3) Simulated nuclei of HL60 cells(Fluo-N3DH-SIM+), and (4) Simulated GFP-actin-stained A549 Lung Cancer cells (Fluo-C3DH-A549-SIM). Note that the labels of the official testing data have not been released. Since each cohort has two training videos, we used one as training while another as testing. The source code of benchmarks (LEID-NL~\cite{dzyubachykleid}, KTH-SE~\cite{magnusson2016segmentation}, and RSHN~\cite{payer2019segmenting}) was also deployed on such data directly. Therefore, the reported results could be different from the online leader board~\cite{isiblb2021}. 

All computation and training were performed via a standard NC6~\cite{azure2020} virtual machine platform at the Microsoft Azure cloud. The virtual machine includes half an NVIDIA Tesla K80 accelerator~\cite{gpu2015} card (12 GB accessible) and six Intel Xeon E5-2690 v3 (Haswell) processor.  The multi-stream training and 3-D Synchronization Algorithm was implemented with tensorflow and Python3. All the models in this study were trained with 20,000 iterations. During model training, the learning rate was initially set to 0.0001, and decreases to 0.00001 after 10,000 iterations. The bandwidth hyper-parameter in mean-shift clustering is set to 0.1.

\begin{table}[t]
\caption{\textbf{Quantitative Result of Empirical Validation}}
\setlength{\tabcolsep}{3pt}
\arrayrulecolor{black}
\renewcommand\arraystretch{1.15}
\centering
\begin{tabular}{p{2.8cm}<{\centering} | p{1.3cm}<{\centering} p{1.3cm}<{\centering} p{1.3cm}<{\centering} | p{1.3cm}<{\centering} p{1.3cm}<{\centering} p{1.3cm}<{\centering} }
\toprule
\multirow{3}{*}{\textbf{Method}} & \multicolumn{6}{c}{\textbf{Data-Set}} \\
\cline{2-7}
& \multicolumn{3}{c|}{Fluo-N3DH-SIM+} & \multicolumn{3}{c}{Fluo-C3DH-A549-SIM} \\
\cline{2-7}
& SEG & TRA & OP & SEG & TRA & OP \\
\hline
LEID-NL~\cite{dzyubachykleid} & 0.643\rlap{$^{(3)}$}  & \textbf{0.971}\rlap{$^{(1)}$} & 0.807\rlap{$^{(3)}$} & 0.827\rlap{$^{(2)}$} & 1.000 & 0.914\rlap{$^{(2)}$} \\
KTH-SE~\cite{magnusson2016segmentation}  & 0.774\rlap{$^{(1)}$}  & 0.958\rlap{$^{(2)}$} & 0.866\rlap{$^{(1)}$} & 0.842\rlap{$^{(1)}$} & 1.000 & 0.921\rlap{$^{(1)}$} \\
RSHN~\cite{payer2019segmenting} & 0.748  & 0.909 & 0.829 & 0.798 & 1.000 & 0.899 \\
\midrule
VoxelEmbed (ours)& \textbf{0.818}  & 0.933 & 0.876 & \textbf{0.852} & 1.000 & \textbf{0.926} \\
VoxelEmbed-D (ours) & 0.806  & 0.952 & \textbf{0.879} & 0.838 & 1.000 & 0.919  \\
\bottomrule
\multicolumn{7}{p{12cm}}{* ``(1),(2),(3)" indicates the current ranking in the Cell Tracking Challenge leader board~\cite{isiblb2021}.}
\end{tabular}
\label{tab1}
\end{table}

\begin{table}[tb]
\caption{\textbf{Quantitative Result of Empirical Validation}}
\setlength{\tabcolsep}{3pt}
\arrayrulecolor{black}
\renewcommand\arraystretch{1.15}
\centering
\begin{tabular}{p{2.8cm}<{\centering} | p{1.3cm}<{\centering} p{1.3cm}<{\centering} p{1.3cm}<{\centering} | p{1.3cm}<{\centering} p{1.3cm}<{\centering} p{1.3cm}<{\centering} }
\toprule
\multirow{3}{*}{\textbf{Method}} & \multicolumn{6}{c}{\textbf{Data-Set}} \\
\cline{2-7}
& \multicolumn{3}{p{4cm}<{\centering}|}{Fluo-N3DH-CHO \quad\quad\quad\quad\quad ($\color{red}{20.6\%}$ frames have labels)} &  \multicolumn{3}{p{4cm}<{\centering}}{Fluo-N3DH-CE \quad\quad\quad\quad\quad\quad\quad ($\color{red}{2\%}$ frames have labels)}\\
\cline{2-7}
& SEG & TRA & OP & SEG & TRA & OP  \\
\hline
LEID-NL~\cite{dzyubachykleid} & 0.901\rlap{$^{(4)}$} & 0.923\rlap{$^{(5)}$} & 0.912\rlap{$^{(5)}$} & - & - & - \\
KTH-SE~\cite{magnusson2016segmentation} & \textbf{0.907}\rlap{$^{(2)}$}  & 0.953\rlap{$^{(1)}$} & \textbf{0.930}\rlap{$^{(1)}$} & 0.667\rlap{$^{(3)}$} & \textbf{0.945}\rlap{$^{(1)}$} & 0.806\rlap{$^{(2)}$}  \\
RSHN~\cite{payer2019segmenting} & 0.837 & 0.946 & 0.892 & 0.673 & 0.875 & 0.774 \\
\midrule
VoxelEmbed (ours) & 0.862 & 0.958 & 0.910 & \textbf{0.717} & 0.897 & 0.807 \\
VoxelEmbed-D (ours) & 0.860 & \textbf{0.959} & 0.910 & 0.709 & 0.925 & \textbf{0.817} \\
\bottomrule
\multicolumn{7}{p{12cm}}{* ``(1),(2),(3)" indicates the current ranking of such approach in the Cell Tracking Challenge leader board~\cite{isiblb2021}. LEID did not provide N3DH-CE, the performance is not available(-).}
\end{tabular}
\label{tab2}
\end{table}

\section{Results}
\begin{figure}[t]
\begin{center}
\includegraphics[width=0.88\linewidth]{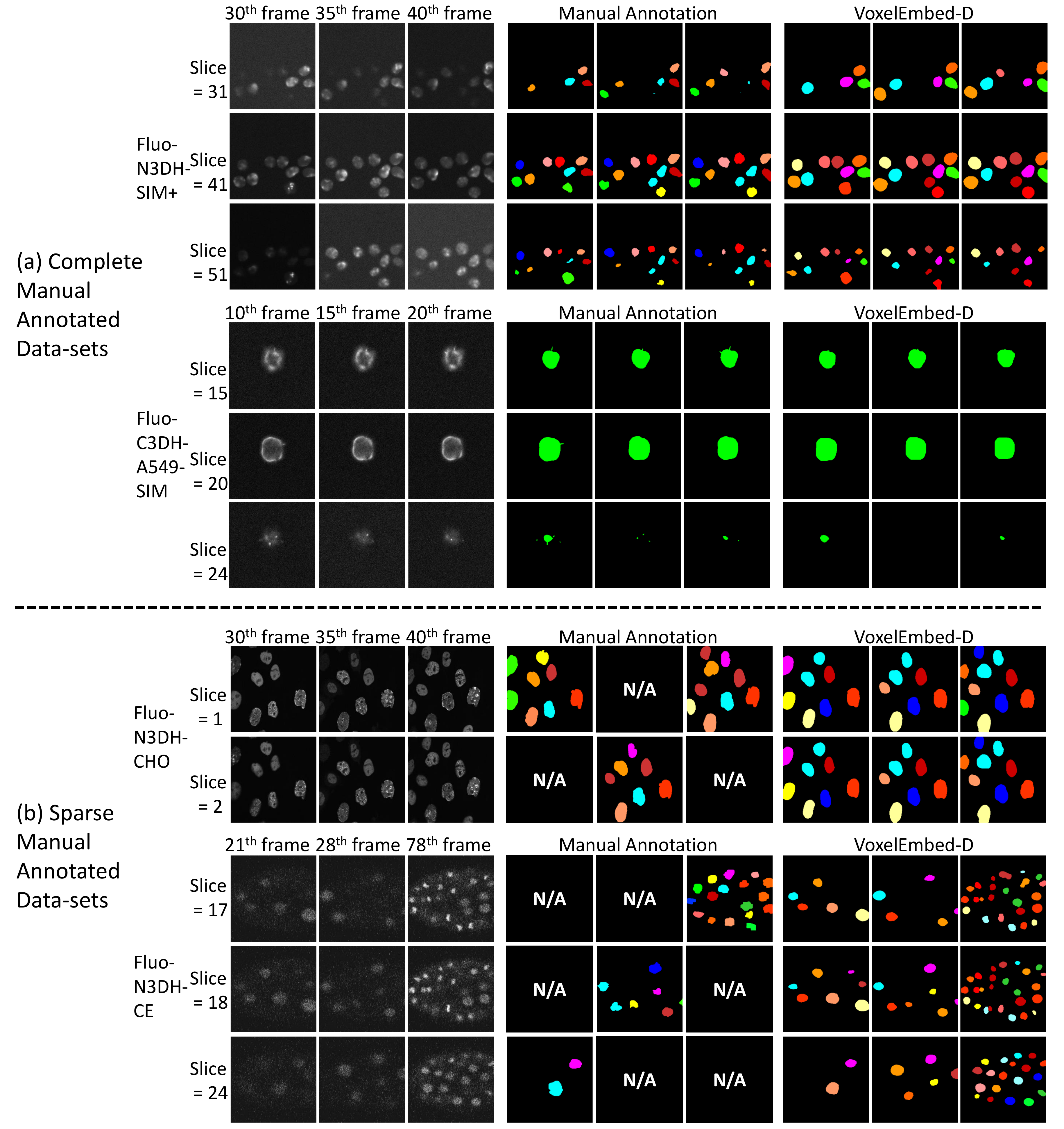}
\end{center}
   \caption{\textbf{The qualitative results of our VoxelEmbed framework}. The 3D ISBI Cell Tracking Challenge dataset, (a) Complete and (b) Sparse manual annotations, as well as the VoxelEmbed results are presented.}
\label{fig5}
\end{figure}

The qualitative results are shown in Fig. \ref{fig5}, while the comparison of quantitative results are presented in Table \ref{tab1} and \ref{tab2}. In Table \ref{tab1} and \ref{tab2}, the Cell Tracking Challenge's official tool of measuring the performance of tracking (TRA) and segmentation (SEG) are employed. TRA is computed by the normalized Acyclic Oriented Graph Matching measure (AOGM)~\cite{matula2015cell} and is used as the tracking accuracy metric. The SEG is computed by Jaccard index. Following the ISBI Cell Tracking Challenge's Benchmark, we also compute the overall performance (OP), which is the average of TRA and SEG.

Based on quantitative and qualitative results, our methods achieved a competitive accuracy performance, using the same network without heavy parameter tuning. For Fluo-N3DH-SIM+ and Fluo-C3DH-A549-SIM data-set, since the manual annotations are available for all video frames, our VoxelEmbed achieved the best SEG and OP. The tracking performance is also competitive. For Fluo-N3DH-CHO and Fluo-N3DH-CE, only sparse manual annotations are provided (Fig. \ref{fig5} (b) ). As a result, the OP of Fluo-N3DH-CHO is inferior compared with leading approaches. 


\section{Conclusion}
In this paper, we introduce VoxelEmbed, an novel embedding based deep learning approach for 3D cell instance segmentation and tracking. The proposed detection method introduced a simple multi-stream training strategy to allow the embedding encoder to learn the spatial-temporal consistent voxel embedding across 3D context. The results show that the VoxelEmbed achieves decent performance compared with the leading method in four Cell Tracking Challenge datasets, with a uniformed learning framework. This study shows the promises of performing 3D cell instance segmentation and tracking with embedding based deep learning with a single GPU.






%
%
\bibliographystyle{splncs04}
\bibliography{main}
%




\end{document}